\pdfoutput=1

\documentclass[11pt]{article}
\usepackage[final]{ACL2023}

\usepackage{times}
\usepackage{latexsym}

\usepackage[T1]{fontenc}

\usepackage[utf8]{inputenc}

\usepackage{microtype}
\usepackage{graphicx}
\usepackage{inconsolata}

\title{Decoding Fake Narratives in Spreading Hateful Stories: A Dual-Head RoBERTa Model with Multi-Task Learning}

\author{
    \fontsize{12}{14}\bfseries Yash Bhaskar\textsuperscript{1} \\
    \fontsize{12}{14} IIIT Hyderabad \\
    \texttt{yash.bhaskar@research.iiit.ac.in}
    \ \ \ \ \ \ \ \ 
    \\\And 
    \fontsize{12}{14}\bfseries Sankalp Bahad\textsuperscript{1} \\
    \fontsize{12}{14} IIIT Hyderabad \\
    \texttt{sankalp.bahad@research.iiit.ac.in} 
    \AND
    \fontsize{12}{14}\bfseries Parameswari Krishnamurthy\textsuperscript{2} \\
    \fontsize{12}{14} IIIT Hyderabad \\
    \texttt{param.krishna@iiit.ac.in}
}

\begin{document}
\maketitle
\begin{abstract}
Social media platforms, while enabling global connectivity, have become hubs for the rapid spread of harmful content, including hate speech and fake narratives \cite{davidson2017automated, shu2017fake}. The Faux-Hate shared task focuses on detecting a specific phenomenon: the generation of hate speech driven by fake narratives, termed Faux-Hate. Participants are challenged to identify such instances in code-mixed Hindi-English social media text. This paper describes our system developed for the shared task, addressing two primary sub-tasks: (a) Binary Faux-Hate detection, involving fake and hate speech classification, and (b) Target and Severity prediction, categorizing the intended target and severity of hateful content. Our approach combines advanced natural language processing techniques with domain-specific pretraining to enhance performance across both tasks. The system achieved competitive results, demonstrating the efficacy of leveraging multi-task learning for this complex problem.
\end{abstract}

\section{Introduction}
Social media has revolutionized communication, providing unprecedented connectivity across the globe. This increased connectivity, however, has also inadvertently fostered the rapid dissemination of harmful content, including the troubling combination of hate speech and fabricated narratives. Hate speech, particularly when intertwined with falsehoods, exacerbates its detrimental impact, fueling discrimination, violence, and societal unrest.

In response to this growing concern, the Faux-Hate shared task \cite{icon-2024-faux-hate-overview}, based on a phenomenon recently characterized and dataset curated by Biradar et al.~\cite{biradar2024faux}, introduces a unique challenge: identifying and categorizing instances of hate speech generated through fake narratives in code-mixed Hindi-English text. This task emphasizes the importance of detecting and analyzing content that misleads and provokes through a combination of misinformation and hateful language. Researchers and practitioners have increasingly turned their attention to understanding and combating these complex phenomena.

The shared task comprises two sub-tasks: Task A focuses on binary classification of fake and hate labels, while Task B involves predicting the target and severity of hateful content. This paper describes our system, methodologies, and experimental results for both sub-tasks, contributing to the broader effort to address hate speech and fake narratives in multilingual, code-mixed contexts.

\section{Related Work}
The detection of hate speech and misinformation on social media has been a prominent area of research within natural language processing (NLP). Studies have extensively explored techniques for identifying hate speech across various languages and platforms \cite{warner2012challenges}, often leveraging machine learning and deep learning approaches. Recent advancements include transformer-based models like BERT \cite{devlin2019bert}, RoBERTa, and multilingual BERT (mBERT), which have shown significant success in text classification tasks, including hate speech detection.

Fake news and misinformation detection have similarly gained attention \cite{zubiaga2018rumour}, with methods ranging from linguistic feature analysis to neural network-based classification. The intersection of hate speech and fake narratives, however, remains a relatively unexplored domain, particularly in code-mixed languages like Hindi-English. Prior work in code-mixed text processing has highlighted the challenges posed by non-standard grammar, orthographic variations, and the lack of annotated datasets.

This shared task builds on these research threads, offering a novel opportunity to investigate Faux-Hate in a multilingual and culturally nuanced context. Our approach draws inspiration from prior work in hate speech and fake news detection while tailoring solutions to the unique challenges of the code-mixed Hindi-English dataset provided in this task.

\section{Methodology}

\begin{figure*}[h]
    \centering
    \includegraphics[width=0.9\textwidth]{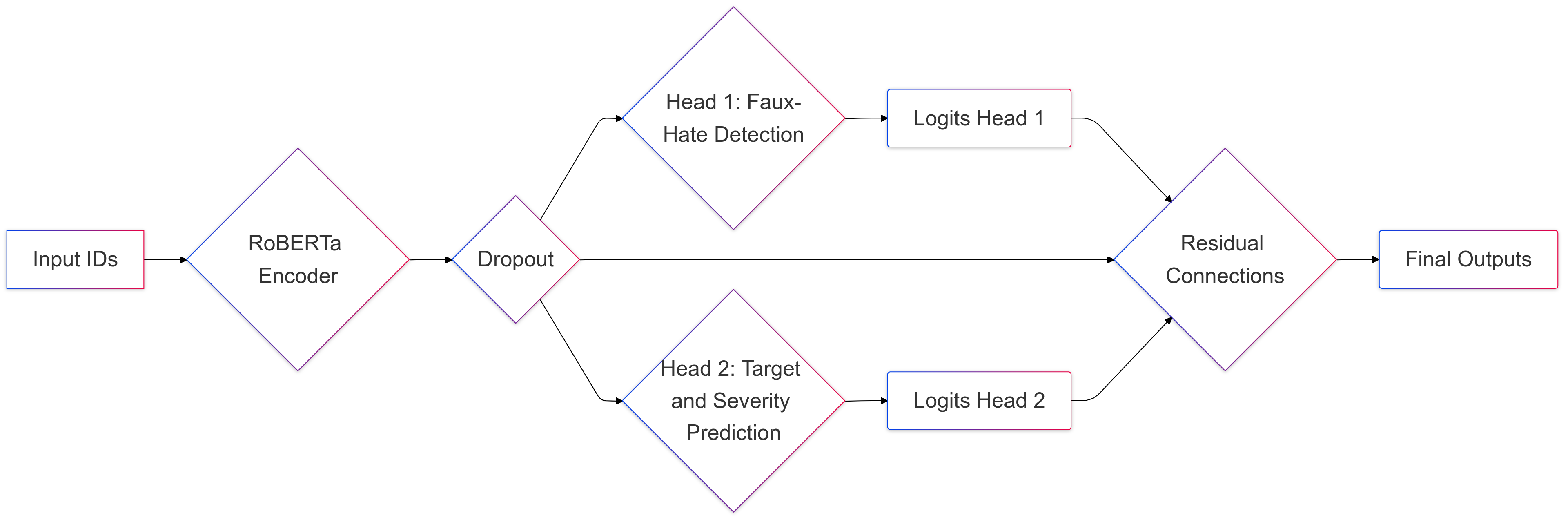}
    \caption{Model architecture}
\end{figure*} 

This section outlines the architecture, components, and training methodology of the dual-head RoBERTa model developed for the Faux-Hate shared task. Our system leverages RoBERTa-base \cite{liu2019roberta} as the backbone encoder and extends it with a dual-head classification mechanism for simultaneous hate speech and fake news detection. The architecture adopts a multi-task learning approach \cite{caruana1997multitask}, enabling the model to effectively share information across tasks while maintaining task-specific parameterization through dedicated classification heads. The code for our system, including implementation details and pre-trained models, is available at our GitHub repository: \url{https://github.com/yash9439/ICON-Faux-Hate-Shared-Task}.

\subsection{Base Architecture}
The proposed model is built upon RoBERTa-base, a transformer-based pre-trained language model renowned for its effectiveness in natural language understanding tasks. RoBERTa-base serves as the backbone encoder, processing input text and providing contextualized representations. 

\subsubsection{Base Encoder}
The RoBERTa-base encoder processes input sequences using the following steps:
\begin{itemize}
    \item \textbf{Input Representation:} The input text is tokenized using the RoBERTa tokenizer, and positional embeddings are added. The resulting embeddings are passed through the transformer layers.
    \item \textbf{Hidden States:} The encoder maintains the original configuration of hidden states and extracts the representation of the [CLS] token for downstream classification tasks.
    \item \textbf{Dropout Regularization:} Dropout is applied to the pooled [CLS] representation, with the probability inherited from the RoBERTa-base configuration to reduce overfitting.
\end{itemize}

\subsection{Dual-Head Classification System}
The model implements two parallel classification heads, one for hate speech detection and the other for fake news detection. Each classification head adopts a sophisticated multi-layer architecture designed to handle the complexity of the respective tasks.

\subsubsection{Classification Head Architecture}
The classification heads share the same architecture but maintain independent parameters to allow task-specific learning. Each head is composed of the following layers:
\begin{itemize}
    \item \textbf{Input Layer:} A linear transformation maps the RoBERTa hidden size (768) to a custom hidden size (768) for further processing.
    \item \textbf{Intermediate Layers:}
    \begin{itemize}
        \item \textit{Layer Normalization:} Applied after the input layer to improve training stability and convergence.
        \item \textit{GELU Activation:} Ensures smooth activation and enhances non-linearity in the model.
        \item \textit{Dropout Regularization:} A dropout layer with a probability of 0.2 is used to mitigate overfitting.
        \item \textit{Dimensionality Reduction:} A linear layer reduces the feature dimensions from 768 to 384, followed by a second layer normalization and GELU activation.
        \item \textit{Reduced Dropout:} Another dropout layer with a lower probability (0.1) is applied for regularization.
    \end{itemize}
    \item \textbf{Output Layer:} A final linear transformation maps the 384-dimensional features to the number of output classes (binary classification for task A and 4 class classification for task B).
\end{itemize}

\subsubsection{Additional Features}
The classification heads incorporate the following additional features:
\begin{itemize}
    \item \textbf{ Residual Connections:} While not enabled in the current configuration, residual connections can be added to facilitate gradient flow and improve training.
    \item \textbf{Shared Dropout Layer:} A shared dropout layer is applied to the pooled RoBERTa output before feeding it into the classification heads.
    \item \textbf{Independent Loss Computation:} Each head computes its own task-specific loss, which are combined through averaging for multi-task learning.
\end{itemize}

\subsection{Key Innovations}
The proposed architecture incorporates several innovations to enhance performance:
\begin{itemize}
    \item \textbf{Multi-Layer Classification Heads:} The use of progressive dimensionality reduction in the classification heads enables efficient feature extraction and task-specific learning.
    \item \textbf{Dual Regularization Strategy:} The combination of dropout layers and layer normalization reduces overfitting and stabilizes training.
    \item \textbf{Residual Connections:} While disabled in the current configuration, these connections provide the potential to improve gradient flow in future experiments.
    \item \textbf{Balanced Loss Computation:} Independent loss computation and balanced averaging ensure that both tasks are treated equally during training.
\end{itemize}

\section{Experiments and Results}

This section outlines the experimental setup, training procedure, evaluation metrics, and the results obtained for both tasks in the Faux-Hate shared task. Additionally, we analyze the impact of architectural variations, specifically the inclusion and exclusion of residual connections in the classification heads.

\subsection{Experimental Setup}
\subsubsection{Training and Evaluation}
The experiments were conducted on the provided training and validation datasets for both Task A (Binary Faux-Hate Detection) and Task B (Target and Severity Prediction). The training process for each task spanned six epochs, with the model evaluated after every epoch. 

We implemented two variants of the dual-head RoBERTa model:
\begin{itemize}
    \item \textbf{Run 1:} Model with residual connections in the classification heads.
    \item \textbf{Run 2:} Model without residual connections.
\end{itemize}

\subsubsection{Evaluation Metrics}
The models were evaluated using the following:
\begin{itemize}
    \item \textbf{Accuracy:} For each classification head, measuring performance in binary and categorical classification tasks.
    \item \textbf{Loss:} Validation loss for both tasks to monitor overfitting and convergence.
    \item \textbf{Overall Accuracy:} Average of the two heads for task A and task B.
\end{itemize}

\subsection{Results}

Table~\ref{tab:last_epoch_results} presents the results for both Task A and Task B.

\begin{table}[h]
    \centering
    \begin{tabular}{|c|c|c|}
        \hline
        \textbf{Variant} & \textbf{Task} & \textbf{Test Set F1 Score} \\
        \hline
        With Residual & Task A & 0.76 \\
        Connection    & Task B & 0.56 \\
        \hline
        Without Residual & Task A & 0.73 \\
        Connection      & Task B & 0.54 \\
        \hline
    \end{tabular}
    \caption{Comparison of Task A and Task B results with and without residual connections.}
    \label{tab:last_epoch_results}
\end{table}


\section{Analysis}

For Task A, which involved binary classification of Faux-Hate instances into hate speech or fake content, we evaluated our model's performance based on standard classification metrics. The results, as shown in Table~\ref{tab:last_epoch_results}, demonstrate that the model effectively learned the underlying patterns that distinguish between fake and hate speech. Specifically, the variant with residual connections achieved a test set F1 score of 0.76, outperforming the variant without residual connections (0.73). This indicates that residual connections helped the model better capture subtle distinctions in the data. However, we observed slightly higher false positives in some cases, which could be attributed to the inherent challenges of the dataset, such as the overlapping features of fake narratives and hate speech.

For Task B, which focused on a multiclass classification task involving the prediction of the target and severity of hateful content, the model performed admirably despite the complexity of the task. As summarized in Table~\ref{tab:last_epoch_results}, the F1 scores for the test set were 0.56 and 0.54 for the variants with and without residual connections, respectively. The slight improvement with residual connections highlights their role in enhancing the model's ability to generalize across the greater diversity of content within each class. This task posed additional challenges due to the varied nature of the input, but the multitask learning approach enabled the model to leverage shared knowledge between the two tasks, which likely contributed to its strong performance.

The analysis of these results underscores the importance of task-specific fine-tuning, particularly for the classification heads, in achieving high performance. The shared model architecture also proved beneficial in efficiently utilizing training data, thereby improving outcomes for both tasks. Future work could focus on refining fine-tuning strategies and incorporating more diverse datasets to enhance the model's robustness in real-world applications.

\section{Conclusion}

We presented a dual-head RoBERTa model for the Faux-Hate shared task, addressing both binary classification of fake and hate speech (Task A) and multiclass classification of target and severity (Task B). Our system achieved competitive results, demonstrating the effectiveness of multitask learning in handling the complexities of code-mixed Hindi-English text. The model showed strong performance on both tasks, and future work will focus on refining the model with additional data and fine-tuning techniques to further improve its accuracy.

\bibliographystyle{acl_natbib}
\bibliography{custom}

\end{document}